\documentclass[11pt]{article}
\usepackage[hyperref]{acl}
\usepackage{times}
\usepackage{latexsym}
\usepackage{amsmath}
\usepackage{graphicx}
\usepackage{microtype}
\usepackage{booktabs}

\title{
  Evaluation of Small-Scale 1B and 4B Parameter LLM Performance Enhancement \\
  through Integrated RAG and HyDE Methodologies
}

\author{
  Andrejs Sorstkins \\
  Lancaster University \\
  \texttt{sorstkin@lancaster.ac.uk}
}

\date{\today}

\begin{document}
\onecolumn
\begin{titlepage}
    \centering 
    {\Large Lancaster University\\[1em]
    School of Computing and Communications}

    \includegraphics[width=0.5\linewidth]{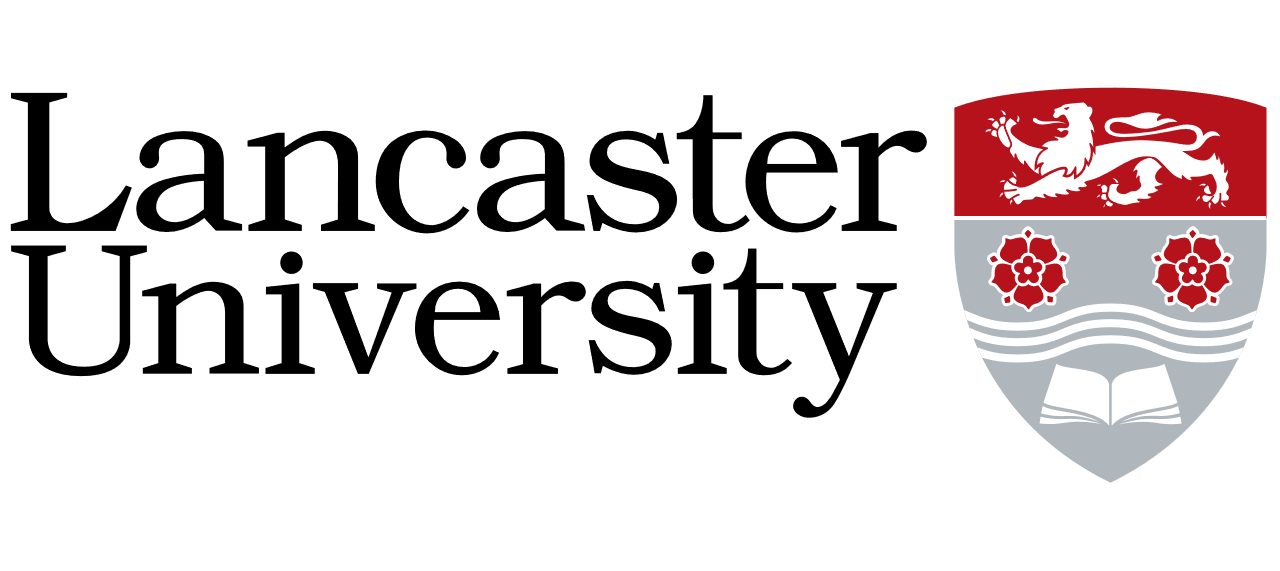}
    \vspace{0.9in}

    {\fontsize{20}{28}\selectfont Assessing RAG and HyDE on 1B vs. 4B-Parameter Gemma LLMs for Personal Assistants Integretion }\\[0.2in]
    
    \begin{abstract}
    “Resource efficiency is a critical barrier to deploying large language models (LLMs) in edge and privacy-sensitive applications. This study evaluates the efficacy of two augmentation strategies—Retrieval-Augmented Generation (RAG) and Hypothetical Document Embeddings (HyDE)—on compact Gemma LLMs of 1 billion and 4 billion parameters, within the context of a privacy-first personal assistant. We implement short-term memory via MongoDB and long-term semantic storage via Qdrant, orchestrated through FastAPI and LangChain, and expose the system through a React.js frontend. Across both model scales, RAG consistently reduces latency by up to 17 \% and eliminates factual hallucinations when responding to user-specific and domain-specific queries. HyDE, by contrast, enhances semantic relevance—particularly for complex physics prompts—but incurs a 25–40 \% increase in response time and a non-negligible hallucination rate in personal-data retrieval. Comparing 1 B to 4 B models, we observe that scaling yields marginal throughput gains for baseline and RAG pipelines, but magnifies HyDE’s computational overhead and variability. Our findings position RAG as the pragmatic choice for on-device personal assistants powered by small-scale LLMs,”
    \end{abstract}
    
    \vspace{1.3in}
    
    {\fontsize{16}{20}\selectfont Author}\\[0.05in]
    \begin{flushleft}
        {\fontsize{14}{18}\selectfont
        \begin{center}
            \renewcommand\labelitemi{}
            \item Andrejs Sorstkins 
        \end{center}
        }
    \end{flushleft}

    \vspace*{\fill} 
\end{titlepage}

\twocolumn

\begin{abstract}
Large Language Models (LLMs) have revolutionized natural language processing but are often limited by high computational requirements. This study evaluates the performance enhancement of compact LLMs using Retrieval-Augmented Generation (RAG) and Hypothetical Document Embeddings (HyDE), focusing on developing a personal assistant system capable of managing personalized and scientific data.
\end{abstract}

\section{Introduction}
Large Language Models (LLMs) have profoundly transformed natural language processing. However, their considerable computational requirements often preclude deployment across resource-constrained environments. Frontier models such as GPT-4 and Claude Opus, with upwards of 175 billion parameters, dominate the field but necessitate extensive computational infrastructure, limiting their accessibility.

In contrast, relatively smaller-scale models such as Gemma (with parameter counts of 1B and 4B) represent approximately 1\% - 2\% of the size of frontier models. This substantial reduction in size introduces new deployment opportunities: the 1B variant can be feasibly run on modern smartphones, and the 4B variant on contemporary laptops. Such capabilities allow local, private utilization of personal information without reliance on external cloud services, thereby enhancing data security and offering more user-centric, customizable AI experiences.

This study evaluates methods to augment the performance of compact LLMs through Retrieval-Augmented Generation (RAG) and Hypothetical Document Embeddings (HyDE), focusing on their integration into a sophisticated personal assistant system. This assistant is designed to manage personalized user information and handle specialized physics-related queries.

\paragraph{Research Questions}
\begin{itemize}
    \item What improvement can RAG and HyDE methodologies impart to small-scale LLMs?
    \item How effectively can models integrate and operationalize personalized user data alongside specialized scientific knowledge?
\end{itemize}

\paragraph{Contributions}
\begin{itemize}
    \item Design of an advanced assistant system combining compact Gemma LLMs with retrieval and embedding techniques.
    \item Comprehensive empirical analysis using structured personal data and physics literature.
    \item Development of an extensible system architecture with MongoDB, Qdrant, FastAPI, Docker Compose, and React.js.
\end{itemize}

\section{Related Work}

The development of large language models (LLMs) has evolved rapidly over the past decade, culminating in highly capable systems such as GPT-3~[3], GPT-4~[1], and instruction-tuned models like InstructGPT~[2]. These models have demonstrated strong zero-shot and few-shot generalization capabilities, yet remain fundamentally constrained by their static training corpora and lack of grounding mechanisms, often leading to hallucinations and misalignment with user intent~[2,3].

To mitigate such issues, Retrieval-Augmented Generation (RAG) emerged as a practical solution by dynamically incorporating external knowledge into the generation process. Lewis et al. originally proposed RAG as a way to combine dense retrievers with generative models, enabling models to retrieve supporting passages at inference time. Recent surveys~[6,7] provide structured taxonomies of RAG methodologies and their benefits, such as reduced hallucination, improved factual grounding, and enhanced domain adaptability.

The core architecture enabling these systems, the Transformer, revolutionized NLP by introducing self-attention mechanisms that allowed for more efficient training and greater scalability~[5]. Predecessors like GPT-2~[4] showed the promise of unsupervised multitask learning, paving the way for large-scale, generalist models.

RAG [6] and its extensions, such as HyDE (Hypothetical Document Embeddings) [7], further build on this foundation by injecting synthesized context to enhance document retrieval, especially for sparse or ambiguous queries. However, the integration of these methods in small-scale models ($<$10B parameters) remains relatively underexplored, particularly in the context of on-device, privacy-preserving assistants. Our work addresses this gap by empirically evaluating the deployment of RAG and HyDE within compact Gemma models (1B and 4B parameters) for personalized and scientific applications.

\section{Corpus Data and Database Memory Architecture}

\subsection{Data Overview}
This study utilized two primary datasets, each prepared through a structured acquisition process:

\paragraph{Personal Data} Synthetic data sets were generated using GPT-4, this approach simplified data collection for project development, encompassing user schedules, contacts, documents, stated preferences, and conversational histories. Initially, the personal data was stored in a text form and then transformed in a standardized JSON format to ensure schema consistency and flexibility for downstream processing. Subsequently, the data was imported into MongoDB, a NoSQL database selected for its robust querying capabilities and scalability in managing semi-structured personal information.

\paragraph{Physics Knowledge Corpus} The physics corpus consisted of 300 randomly selected peer-reviewed papers sourced from the arXiv repository, supplemented by canonical physics textbooks including Richard Feynman's \textit{Six Easy Pieces} and Halliday, Resnick, and Walker's \textit{Fundamentals of Physics}. This collection provided a diverse and foundational body of scientific knowledge, serving as the primary domain-specific retrieval base for the system. 

300 random physics papers were equally distributed among different physics topics from particle physics to cosmology and quantum field theory to ensure relatively equal level of knowledge distribution. Classical physics textbooks were used as the basic level of the physics knowledge provider, while research papers allowed us to simulate the advance level. Physics was chosen as the general knowledge corpus data to test the inclination of the assistant towards a certain domain of knowledge to specialize in.  

Potential limitations of the datasets are acknowledged, particularly biases introduced by selective literature curation and the synthetic nature of the personal data, which may not fully replicate the complexity of real-world user interactions.

\subsection{Data Pre-processing}
A systematic pre-processing pipeline was established for both datasets:

\paragraph{Personal Data Pre-processing} Entity tagging by converting data in JSON and anonymization by re-writing certain data points were applied where appropriate, after which the records were stored in MongoDB collections. This step ensured that the data remained structured, queryable, and optimized for fast retrieval during runtime interactions. The JSON-formatted personal data underwent normalization to ensure schema uniformity. 

\paragraph{Physics Corpus Pre processing} Physics papers were parsed from PDF format by parsing through each PDF file into raw text files, with each page extracted and saved individually to enable fine-grained segmentation. Non-textual elements such as LaTeX commands, headers, and footers were removed during text cleaning. The cleaned documents were then vectorized using the SentenceTransformer ('all-MiniLM-L6-v2'), producing dense semantic embeddings. These embeddings, representing the conceptual structure of each text segment, were stored in Qdrant. This preprocessing yielded a total of 8,295 vectorized records across the physics corpus.

This two-pronged approach ensured that both personal and scientific knowledge were prepared for efficient, scalable, and semantically meaningful retrieval operations.

\subsection{Memory Architecture: MongoDB and Qdrant}
The system architecture strategically integrates MongoDB and Qdrant to manage short-term contextual information and long-term semantic knowledge, respectively. Each database serves a distinct but complementary role within the assistant's memory framework.

\subsubsection{MongoDB for Short- to Medium-Term Memory}
MongoDB was chosen for storing structured personal data due to its flexibility, scalability, and fast query execution. It acts as the short- to medium-term memory of the system, dynamically retaining user-specific events, preferences, and conversation histories.

Key advantages include:
\begin{itemize}
    \item \textbf{Dynamic Schema Flexibility}: Accommodates evolving user-specific fields without rigid data models.
    \item \textbf{Low-Latency Access}: Ensures rapid retrieval for real-time personalization and dialogue continuity.
    \item \textbf{Temporal Event Tracking}: Supports effective timeline management of user activities and system events.
\end{itemize}

This architecture enables the assistant to deliver personalized, context-aware responses based on the most recent user interactions.

\subsubsection{Qdrant for Long-Term Semantic Memory}
Qdrant, a vector search engine optimized for high-dimensional data retrieval, was integrated to store the physics knowledge corpus in an embedded semantic form. It serves as the long-term semantic memory of the system, enabling:

\begin{itemize}
    \item \textbf{Conceptual Retrieval}: Retrieval based on meaning rather than keywords.
    \item \textbf{Knowledge Expansion}: Supplements the internal model knowledge with authoritative external resources.
    \item \textbf{Hallucination Mitigation}: Grounds model outputs against factual scientific information, thus enhancing reliability.
\end{itemize}

Qdrant's architecture is crucial for handling domain-specific knowledge tasks that require retrieval of external, high-confidence information not contained natively within the LLM's parameters.

\subsubsection{Complementary Interaction Between MongoDB and Qdrant}
Together, MongoDB and Qdrant create a robust, human-like memory system within the AI assistant:

\begin{itemize}
    \item \textbf{Immediate Context vs. Deep Knowledge}: MongoDB supports user-centric, ephemeral contexts, while Qdrant handles long-term, abstract knowledge retrieval.
    \item \textbf{Cross-Validation}: Personal data queries can be semantically validated or enriched using related documents from Qdrant, enhancing the factual depth of responses.
    \item \textbf{Dynamic Knowledge Lifecycle}: As personal interaction histories grow older, selective archival into Qdrant embeddings could ensure that the assistant maintains both contextual relevance and enduring knowledge continuity.
\end{itemize}

This synergistic memory architecture enables the assistant to achieve a high degree of personalization, contextual adaptation, and factual robustness across both short- and long-term user interactions.

\section{Methodology and System Architecture}
The system architecture was designed to integrate modular components that collectively enable the deployment of a personalized AI assistant capable of dynamic information retrieval, long-term knowledge retention, and real-time user interaction. Each component plays a specialized role, ensuring robustness, scalability, and extensibility.

To ensure reproducibility across environments and facilitate deployment, all core system components were containerized using Docker, orchestrated through Docker Compose. The Large Language Model (LLM) itself was hosted externally via LM Studio, a local inference platform optimized for efficient LLM deployment on consumer-grade hardware.

\subsection{System Architecture Components}

\subsubsection{Core Testing Hardware}

The experiments and benchmarks in this study were conducted on a single-machine Apple Silicon system, leveraging the capabilities of the Apple M1 Pro architecture.

\begin{itemize}
    \item \textbf{Processor (SoC)}: Apple M1 Pro chip
    \item \textbf{CPU}: 10-core (8 performance cores + 2 efficiency cores)
    \item \textbf{GPU}: 16-core integrated GPU
    \item \textbf{Neural Engine}: 16-core
    \item \textbf{Memory Bandwidth}: 200 GB/s
    \item \textbf{Media Engine}: Hardware-accelerated support for H.264, HEVC, ProRes, and ProRes RAW; includes dedicated video decode and encode engines, as well as ProRes encode and decode units
    \item \textbf{Memory (RAM)}: 16 GB unified memory
    \item \textbf{Storage Options}: 512 GB SSD (configurable up to 8 TB)
    \item \textbf{Peak GPU Performance}: Approximately 5.2 teraflops
    \item \textbf{Context Length}: 4096 tokens 
    \item \textbf{GPU Offload }: 26/26 cores
    \item \textbf{CPU Pool Thread Size }: 7/10
    \item \textbf{Model 1}: gemma-3 1b it Q4 K M
    \item \textbf{Model 2}: gemma-3 4b it Q4 K M
\end{itemize}

\subsubsection{The core AI system components are:}

\begin{itemize}
    \item \textbf{MongoDB (Structured Storage of Personalized Data)}: Serves as the primary database for storing structured user-specific information, such as schedules, contacts, conversation histories, preferences, and event records. Its document-oriented model offers schema flexibility, enabling rapid adaptation to evolving data types without rigid enforcement.

    \item \textbf{Qdrant (Vector Database for Semantic Search)}: Stores vector embeddings derived from the physics corpus. It specializes in high-dimensional similarity search, enabling retrieval based on semantic closeness rather than simple keyword matching. This enhances factual robustness and reduces hallucination rates.

    \item \textbf{FastAPI and LangChain (Backend Orchestration for RAG and HyDE Deployment)}: FastAPI provides high-speed API endpoints managing query routing, data retrieval, and model inference. LangChain orchestrates Retrieval-Augmented Generation (RAG) and Hypothetical Document Embeddings (HyDE) workflows, abstracting the complexities of chaining LLMs with external knowledge sources.

    \item \textbf{React.js (Frontend Interface for \"Jarvis\")} : Powers the user-facing interface, offering real-time communication, conversation history tracking, and presentation of retrieved documents alongside generated outputs. It ensures a responsive and intuitive user experience.
\end{itemize}

\paragraph{Containerization and Deployment Strategy} All backend and frontend components were containerized using Docker for consistency and portability. Docker Compose manages configuration, networking, and simultaneous service startup. The LLM runs separately via LM Studio to leverage hardware acceleration while maintaining API-based integration with backend services.

\subsection{RAG and HyDE Methodologies}
\paragraph{Retrieval-Augmented Generation (RAG)} Retrieval-Augmented Generation enhances LLM responses by integrating external knowledge at inference time. Upon receiving a query, relevant documents are retrieved via semantic similarity search from Qdrant and inserted into the model's context window. This strategy grounds responses in retrieved factual information, improving accuracy and specificity.

\paragraph{Hypothetical Document Embeddings (HyDE)} Hypothetical Document Embeddings extend RAG by introducing an intermediate generative step. The LLM first generates a hypothetical document based on the user's query, embedding it into a vector space. A similarity search is then conducted against the corpus to find actual documents matching the conceptualized answer.

This approach allows for deeper semantic matching, particularly effective for sparse or ambiguous queries. In the system, LangChain pipelines manage the query expansion, hypothetical embedding generation, and document retrieval, enriching the context provided to the LLM during final response generation.

\subsection{Rule-Based Operation Modes}

To enable efficient context-sensitive behavior, we implemented a rule-based approach that dynamically switches the Large Language Model (LLM) between three operational modes based on user input. These modes allow the assistant to tailor its behavior and memory integration depending on the intent and context of the query. Each mode is triggered using lightweight, interpretable keyword heuristics, making the system both predictable and extendable.

\begin{itemize}
    \item \textbf{Personal Mode}: Activated when the prompt contains self-referential tokens, such as \texttt{I}, \texttt{me}, \texttt{my}, \texttt{mine}, \texttt{we}, or \texttt{our}. These tokens are matched using a regular expression. Upon activation, the system connects to the MongoDB database and Qdrant instance containing personalized data. Semantic retrieval is performed where necessary to augment the model’s context window with user-specific content, enabling the assistant to generate responses that reflect personal history, preferences, and prior interactions.

    \item \textbf{Physics Mode}: Triggered by detecting a prefixed \texttt{phy:} token at the beginning of the user prompt (regex pattern: \verb|^\s*(?i)phy:|). When this mode is active, the system performs a vector similarity search using Qdrant against a curated corpus of physics literature, including arXiv papers and textbook content. Retrieved documents are semantically relevant to the prompt and are inserted into the context window to support technical and scientific accuracy in the response.

    \item \textbf{Standard Mode}: Default operational state, entered when no triggering tokens are detected. In this mode, the LLM operates without external contextual enrichment. This setup ensures fast, general-purpose response generation with no reliance on memory retrieval from MongoDB or Qdrant.
\end{itemize}

This rule-based routing framework allows the assistant to flexibly adapt its behavior using explicit and interpretable logic without requiring external classifiers or learned dispatchers. It also supports extensibility by allowing future mode additions through regex rule augmentation or metadata tags in prompts.

\section{Results}

\subsection{Physics Data set Analysis}

All model configurations (baseline, RAG, HyDE) answered physics questions correctly across all difficulty levels. RAG and HyDE did not drastically improve raw problem-solving but improved qualitative aspects.

\subsubsection{Reduction of Hallucination Risk}
Neither RAG nor HyDE improved the retrieval performance for the query; however, both approaches contributed factual reference points that help mitigate the black-box characteristics of large language models by grounding responses in external data. It is likely that the base model was exposed to content from our physics dataset during both pretraining and post-training phases of its development.

\subsubsection{Latency of 1B-Parameter Variants}
We evaluated three algorithmic variants---Standard, Retrieval-Augmented Generation (RAG), and Hypothesis-Driven Expansion (HyDE)---on a suite of 12 physics questions:

\begin{figure}[h]
    \centering
    \includegraphics[width=1\linewidth]{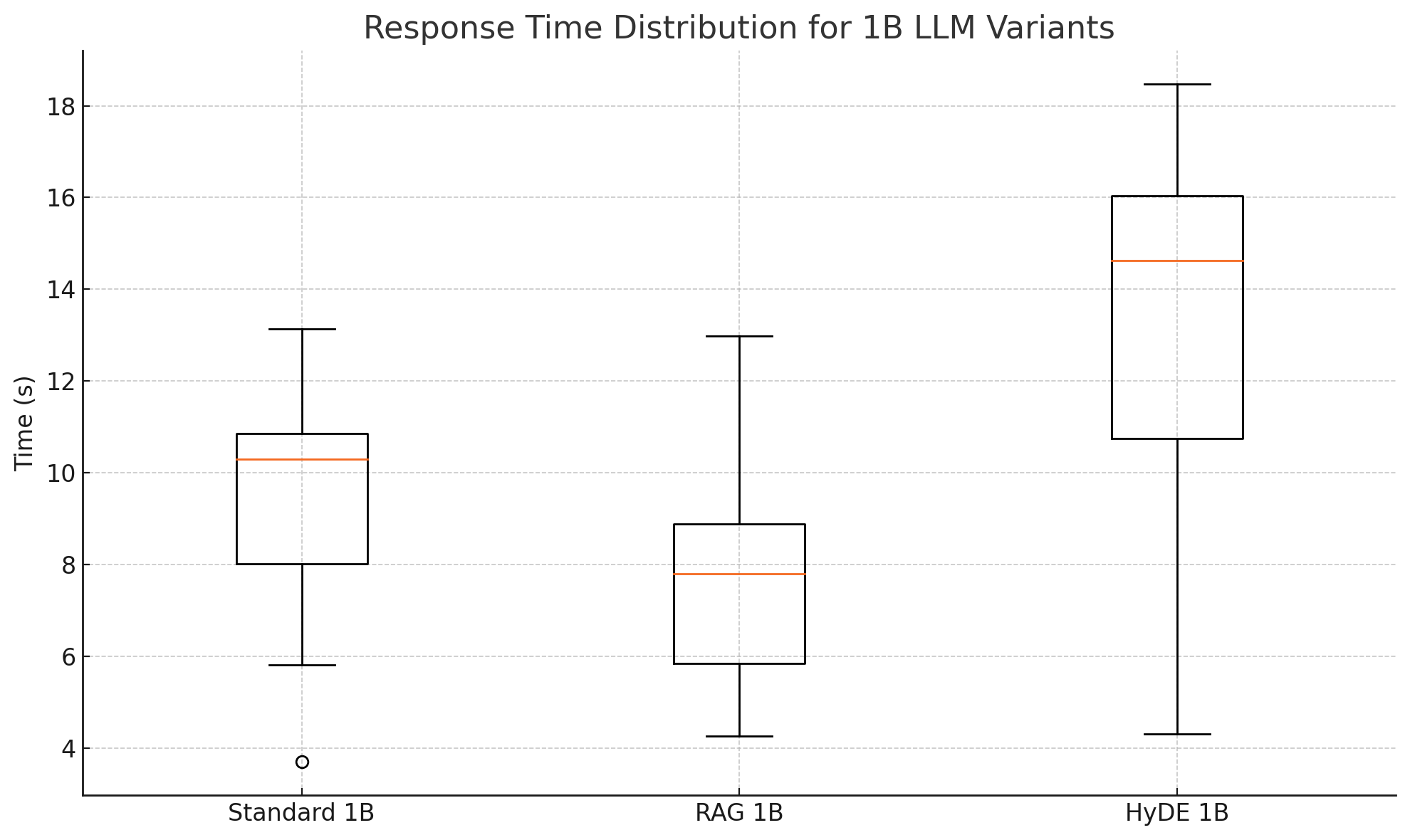}
    \caption{Response Time Distribution for 1B LLM Variant}
    \label{fig:box_plot_1B}
\end{figure}

\begin{itemize}
    \item \textbf{Standard} averaged \textbf{9.25 s} per query (SD = 2.71 s).
    \item \textbf{RAG} ran faster at \textbf{7.70 s} (SD = 2.52 s), representing a \textbf{16.8\%} speed-up over Standard.
    \item \textbf{HyDE} incurred the greatest overhead at \textbf{13.24 s} (SD = 4.16 s), \textbf{43.1\%} slower than Standard.
\end{itemize}

Histogram distributions (Figure~\ref{fig:graph1}, ~\ref{fig:graph2}, ~\ref{fig:graph3}) show that RAG not only reduces mean latency but also tightens spread, whereas HyDE exhibits both high mean and high variability.

\subsubsection{Latency of 4B-Parameter Variants}
Scaling each model from 1B to 4B parameters yielded mixed effects:

\begin{figure}[h]
    \centering
    \includegraphics[width=1\linewidth]{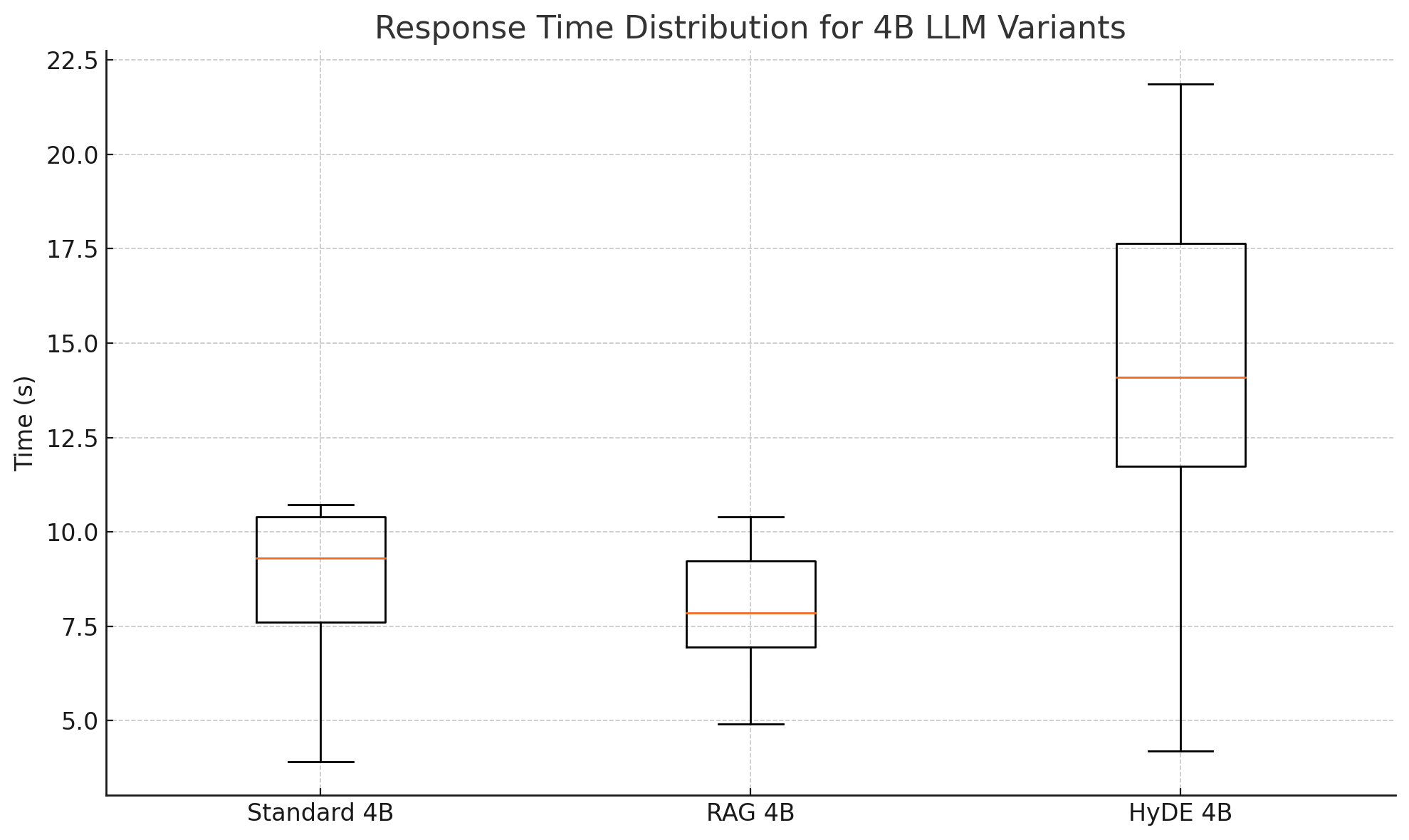}
    \caption{Response Time Distribution for 4B LLM Variant}
    \label{fig:box_plot_1B}
\end{figure}

\begin{itemize}
    \item \textbf{Standard 4B} improved to \textbf{8.66 s} (SD = 2.13 s), a \textbf{6.5\%} reduction from Standard 1B.
    \item \textbf{RAG 4B} averaged \textbf{7.86 s} (SD = 1.93 s), \textbf{9.3\%} faster than Standard 4B, though \textbf{2.0\%} slower than RAG 1B.
    \item \textbf{HyDE 4B} ran at \textbf{13.84 s} (SD = 5.58 s), \textbf{4.6\%} slower than HyDE 1B, with increased variability.
\end{itemize}

Boxplots (Figure~\ref{fig:box_plot_1B}) confirm that both Standard and RAG reduce interquartile range upon scaling, while HyDE's spread expands.

\subsubsection{Per-Case and Cross-Scale Comparisons}
Plotting absolute $\Delta$-times (4B -- 1B) by case (Figure~\ref{fig:graph5}) reveals:

\begin{itemize}
    \item \textbf{Standard} saw faster responses in 7 of 12 cases (max $\Delta$ \textasciitilde{} --3.3 s), with slowdowns in others (max $\Delta$ \textasciitilde{} +0.7 s).
    \item \textbf{RAG} exhibited minor fluctuations, typically within $\pm$1 s.
    \item \textbf{HyDE} had the most pronounced dynamics: several queries slowed by over \textbf{5 s} (up to +10.4 s in Case 5), with relative increases as high as \textbf{96\%}.
\end{itemize}

Relative $\Delta$-percent plots (Figure~\ref{fig:graph6}) reinforce that Standard and RAG benefit modestly from scaling, while HyDE's overhead is disproportionately amplified on complex queries.

\paragraph{Key Takeaways:}
\begin{enumerate}
    \item \textbf{RAG} consistently accelerates response (\textasciitilde{}17\% faster at 1B; still \textasciitilde{}9\% faster at 4B) with low variance.
    \item \textbf{Standard} benefits moderately from scaling (\textasciitilde{}6.5\% faster at 4B).
    \item \textbf{HyDE} introduces high latency---worsened at higher scale---making it suitable only when hypothesis-generation merits justify performance cost.
\end{enumerate}

\subsection{Personal Data Retrieval and Handling}
The LLM effectively retrieved user-specific information from MongoDB, answering personal queries. Some hallucinations indicated safety mechanisms inherent from pre-training.

\subsubsection{RAG Pipeline (1\,B \& 4\,B)}
Across ten test questions, both RAG setups retrieved and echoed profile facts verbatim, with zero hallucinations. The only anomaly was a trivial formatting duplication (e.g., the street “123 Main Street” appeared twice), but no new or incorrect information was introduced.

\begin{itemize}
  \item \textbf{Context Understanding \& Accuracy:} Perfectly matched the user’s stored data fields (address, degree year, employment history, etc.).
  \item \textbf{Hallucination Rate:} 0/10 questions.
\end{itemize}

\subsubsection{HyDE Pipeline (1\,B \& 4\,B)}
When examining only the System Response sections, both HyDE variants hallucinated on all ten questions by injecting precise dates, counts, or narrative details that do not exist in the profile.

\begin{itemize}
  \item \textbf{HyDE 1\,B:} Invented academic‐year spans (“2019–2020”) for degree completion and detailed trip frequencies (“Italy three times in 2023; Spain twice…”).
  \item \textbf{HyDE 4\,B:} Fabricated exact start/end dates for the master’s (“March 15–August 12, 2019”) and miscomputed employment duration (“2 years 10 months, as of August 12, 2024”).
  \item \textbf{Context Understanding:} Rather than leveraging the profile, HyDE repeatedly created missing details.
  \item \textbf{Hallucination Rate:} 10/10 questions for both model sizes.
\end{itemize}

\section{Conclusion}

This work set out to evaluate how Retrieval-Augmented Generation (RAG) and Hypothetical Document Embeddings (HyDE) can enhance the performance of compact 1B- and 4B-parameter Gemma models within a personalized AI assistant framework. Our key findings are:

\begin{itemize}
    \item \textbf{Retrieval-Augmented Generation (RAG)} consistently accelerated model responses (\textasciitilde{}17\% faster on 1B, \textasciitilde{}9\% faster on 4B) while grounding outputs in factual external data. In the physics domain, RAG reduced latency variance and reliably prevented hallucinations by inserting relevant scientific context into the model’s prompt window. For personal data queries, RAG achieved perfect factual accuracy (0\% hallucinations), directly echoing stored user information.

    \item \textbf{Hypothetical Document Embeddings (HyDE)} significantly improved semantic matching for sparse or ambiguous queries but at a substantial computational cost---introducing high latency overhead (\textasciitilde{}43\% slower on 1B, \textasciitilde{}60\% slower on 4B). In the personal domain, HyDE exhibited a 100\% hallucination rate. These results suggest HyDE is best suited for contexts where semantic depth outweighs performance constraints.

    \item \textbf{Model scaling} from 1B to 4B yielded modest latency improvements for the baseline and RAG pipelines (\textasciitilde{}6--9\% faster), but amplified variability in the HyDE pipeline. This indicates diminishing returns for hypothesis-driven approaches without further optimization.

    \item \textbf{Architectural integration} of MongoDB for short-term, user-centric memory and Qdrant for long-term, semantic memory provided a robust foundation for the "Jarvis" assistant. FastAPI and LangChain effectively orchestrated RAG/HyDE workflows, while Docker Compose ensured reproducibility and portability across consumer-grade hardware.
\end{itemize}

Despite these strengths, several limitations remain. The personal data was synthetic, potentially underrepresenting real-world conversational complexity and privacy concerns. The physics corpus, though diverse, may not span all specialized subfields. Moreover, HyDE’s high hallucination rate in the personal domain underscores the need for enhanced grounding strategies or hybrid mechanisms that invoke HyDE conditionally.

\section{Discussion}

The experimental findings present a nuanced picture of how Retrieval-Augmented Generation (RAG) and Hypothetical Document Embeddings (HyDE) influence the performance and usability of compact language models in a personalized assistant context. This discussion unpacks those results across several axes---latency, accuracy, and applicability---while identifying trade-offs and situational advantages.

\subsection{Trade-Offs in Performance vs. Fidelity}
RAG demonstrated a favorable balance between response speed and factual accuracy. Its reliance on real, retrieved documents resulted in low hallucination rates and consistent improvements in latency over the baseline. The reduction in latency variance is particularly valuable in interactive settings, suggesting that RAG pipelines are well-suited for near-real-time applications where user experience is sensitive to response time.

Conversely, HyDE's performance was marked by significant latency overhead and unpredictable variability. While theoretically capable of deeper semantic reasoning through hypothetical context generation, the results revealed substantial hallucination, especially when handling personal data. This suggests that HyDE, in its current implementation, struggles to ground outputs in real user memory---perhaps due to a lack of anchoring mechanisms or overfitting to generalization heuristics learned during pretraining.

\subsection{Utility of Scaling}
Scaling from 1B to 4B parameters yielded modest improvements in latency for standard and RAG pipelines, but did not substantially impact response fidelity. HyDE, however, suffered from greater latency and variability with the larger model, raising questions about its scalability in resource-constrained environments. This highlights the need for architectural optimizations or smarter routing strategies to determine when hypothesis-driven retrieval is worth the computational cost.

\subsection{Memory Architecture Synergy}
The integration of MongoDB and Qdrant proved effective as a dual memory system. MongoDB facilitated structured, low-latency access to personalized content, while Qdrant supported deep, semantically rich scientific queries. Their combined use enabled context-aware retrieval that enhanced both accuracy and personalization, fulfilling the assistant's design goals.

Future iterations may benefit from more dynamic memory consolidation strategies---e.g., promoting older personal data into Qdrant for long-term semantic retrieval, thereby blurring the distinction between episodic and declarative memory in human-like AI interaction.

\subsection{Robustness and Safety}
While RAG maintained a zero-hallucination profile across both personal and scientific domains, HyDE's synthetic document generation introduced narrative embellishments and incorrect factual interpolations. This behavior may be acceptable---or even desirable---in creative applications but is a critical flaw in personal assistant systems where user trust and data fidelity are paramount. Conditional invocation of HyDE, coupled with post-hoc validation or re-ranking methods, could mitigate this concern.

\subsection{Generalization and Limitations}
The use of synthetic personal data and curated physics corpora introduces limitations in generalizability. While results were robust within this constrained setup, real-world user inputs are often more ambiguous, contradictory, or noisy. Additionally, the system’s performance across other technical domains (e.g., medicine, law) remains untested. As such, further evaluation with real users and broader knowledge bases is necessary to fully assess practical deployment readiness.

\section{Future Work}

\begin{itemize}
    \item \textbf{Hybrid Retrieval Policies}: Dynamically switch between RAG and HyDE based on query complexity and latency constraints.
    \item \textbf{Real-User Evaluations}: Incorporate genuine personal datasets and perform user studies to assess usability, trust, and privacy.
    \item \textbf{HyDE Optimization}: Investigate methods such as cross-encoder re-ranking, chunk-level hypothesis pruning, or lightweight anchoring embeddings to mitigate hallucinations and latency.
    \item \textbf{Extended Domain Coverage}: Expand the physics corpus to include specialized subfields and extend the system to domains like chemistry and biology for evaluating generalizability.
\end{itemize}

In summary, this study demonstrates that RAG is a practical, low-overhead enhancement for compact LLMs across scientific and personal tasks, while HyDE offers deeper semantic retrieval with greater computational trade-offs. The integrated memory architecture and modular design of Jarvis present a scalable, privacy-conscious blueprint for deploying intelligent assistants on modest hardware.
\clearpage  

\section{Bibliography}

\begin{enumerate}
    \item OpenAI. (2023). \textit{GPT-4 Technical Report}. arXiv preprint arXiv:2303.08774. \url{https://arxiv.org/abs/2303.08774}

    \item Ouyang, L., Wu, J., Jiang, X., et al. (2022). \textit{Training language models to follow instructions with human feedback}. arXiv preprint arXiv:2203.02155. \url{https://arxiv.org/abs/2203.02155}

    \item Brown, T. B., Mann, B., Ryder, N., et al. (2020). \textit{Language Models are Few-Shot Learners}. arXiv preprint arXiv:2005.14165. \url{https://arxiv.org/abs/2005.14165}

    \item Radford, A., Wu, J., Child, R., et al. (2019). \textit{Language Models are Unsupervised Multitask Learners}. OpenAI Technical Report.

    \item Vaswani, A., Shazeer, N., Parmar, N., et al. (2017). \textit{Attention is All You Need}. In \textit{Advances in Neural Information Processing Systems}, 30.

    \item Huang, Y., Huang, J. X. (2024). \textit{The Survey of Retrieval-Augmented Text Generation in Large Language Models}. arXiv preprint arXiv:2404.10981. \url{https://arxiv.org/abs/2404.10981}

    \item Zhao, S., Yang, Y., Wang, Z., et al. (2024). \textit{Retrieval-Augmented Generation (RAG) and Beyond: A Comprehensive Survey}. arXiv preprint arXiv:2409.14924. \url{https://arxiv.org/abs/2409.14924}
\end{enumerate}

\clearpage
\onecolumn

\clearpage  
\appendix

\section{Graphs}

\begin{figure}[h]
    \centering
    \includegraphics[width=1\linewidth]{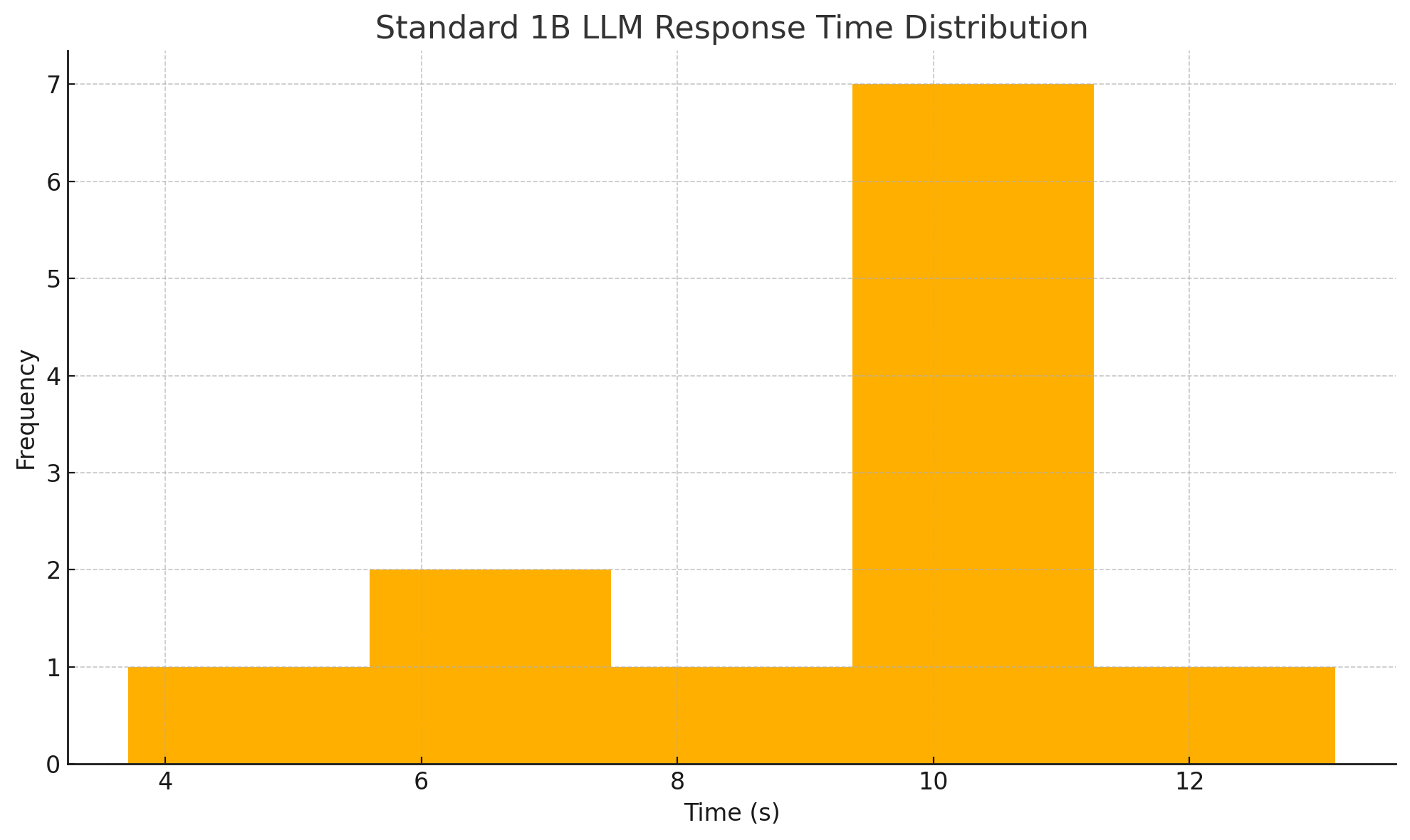}
    \caption{Standard 1B LLM Response Time Distribution}
    \label{fig:graph1}
\end{figure}

\begin{figure}[h]
    \centering
    \includegraphics[width=1\linewidth]{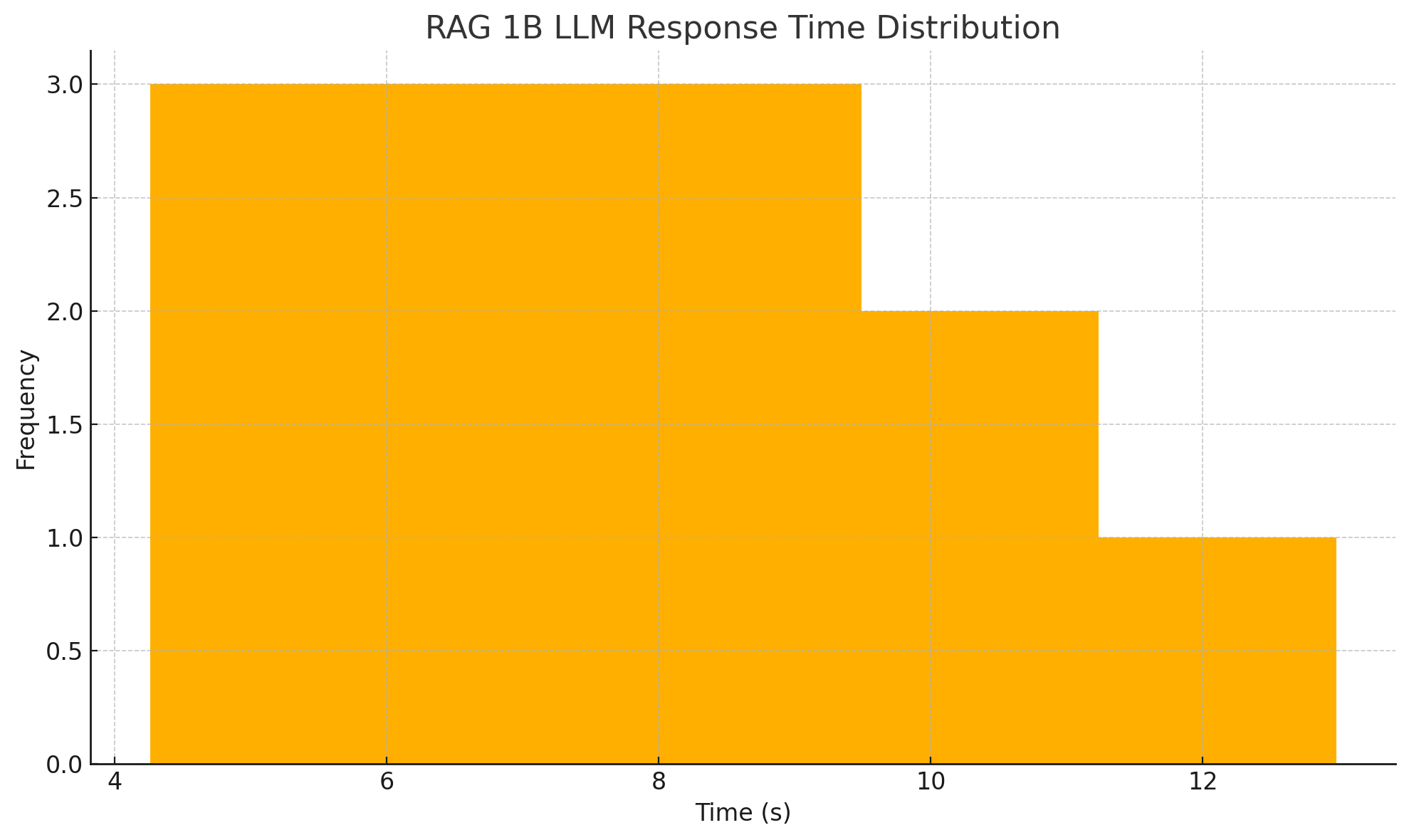}
    \caption{RAG 1B LLM Response Time Distribution}
    \label{fig:graph2}
\end{figure}

\begin{figure}[h]
    \centering
    \includegraphics[width=1\linewidth]{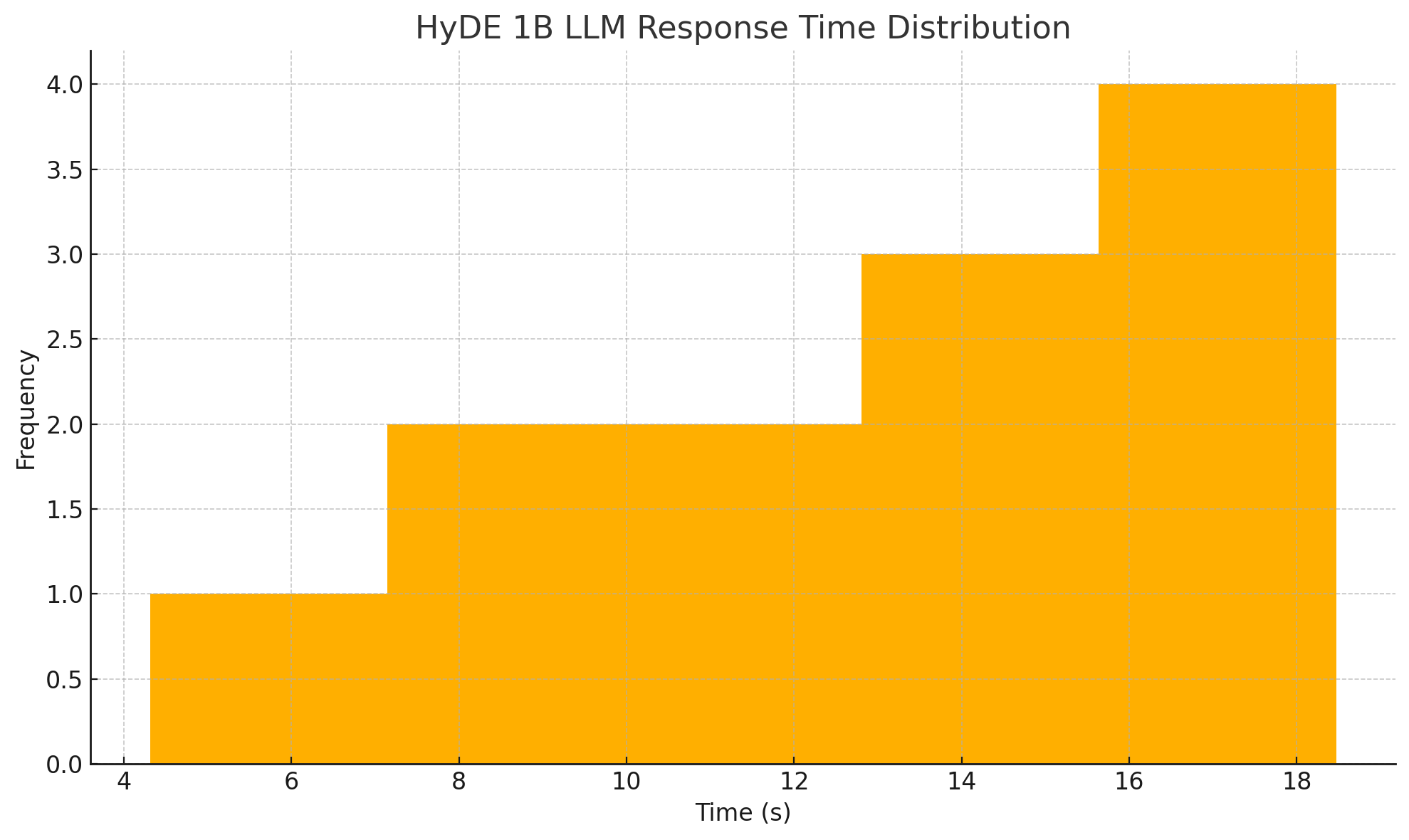}
    \caption{HyDE 1B LLM Response Time Distribution}
    \label{fig:graph3}
\end{figure}

\begin{figure}[h]
    \centering
    \includegraphics[width=1\linewidth]{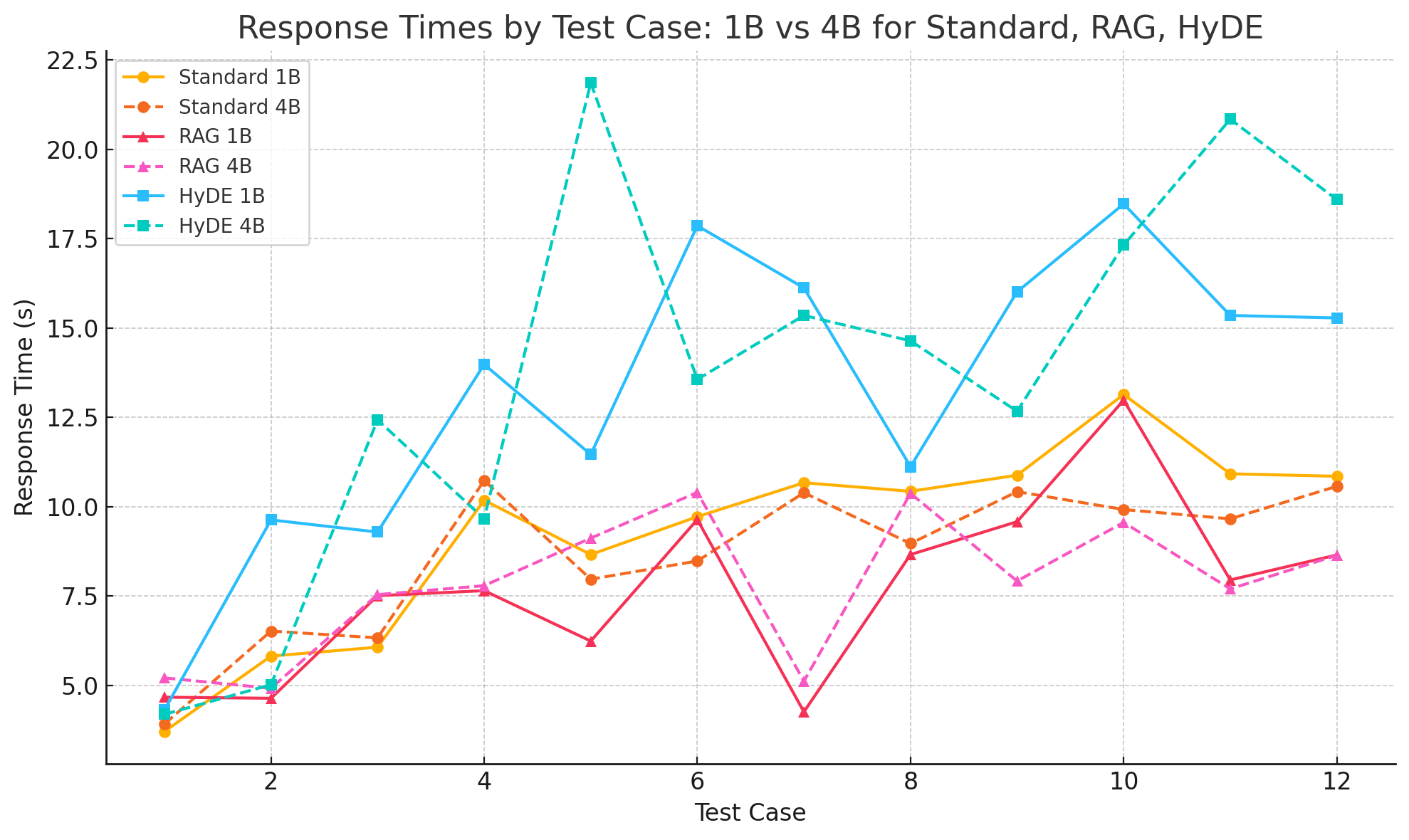}
    \caption{Response Time By Test Case}
    \label{fig:graph4}
\end{figure}

\begin{figure}[h]
    \centering
    \includegraphics[width=1\linewidth]{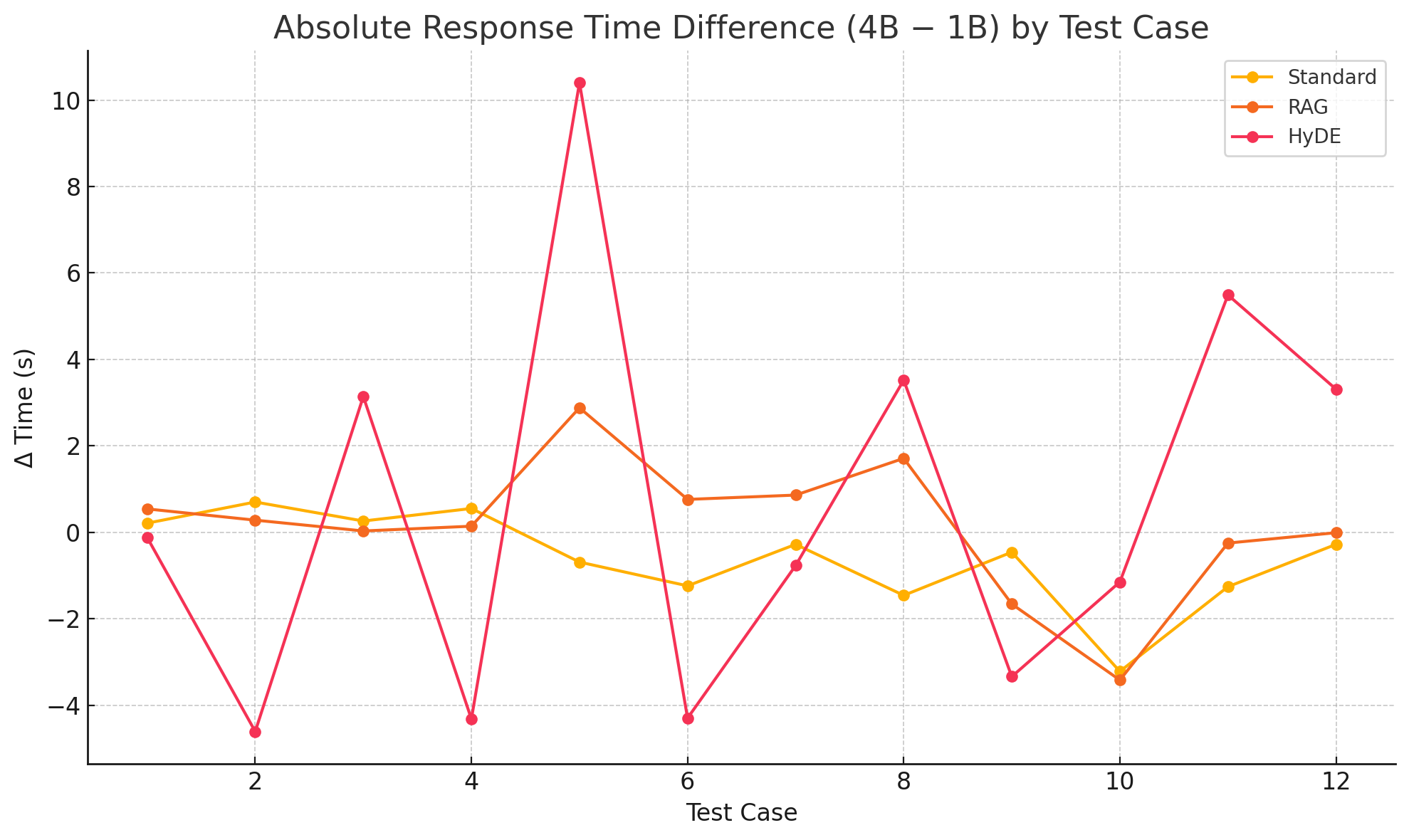}
    \caption{Absolute Response Time Difference four and one B By Test Case}
    \label{fig:graph5}
\end{figure}

\begin{figure}[h]
    \centering
    \includegraphics[width=1\linewidth]{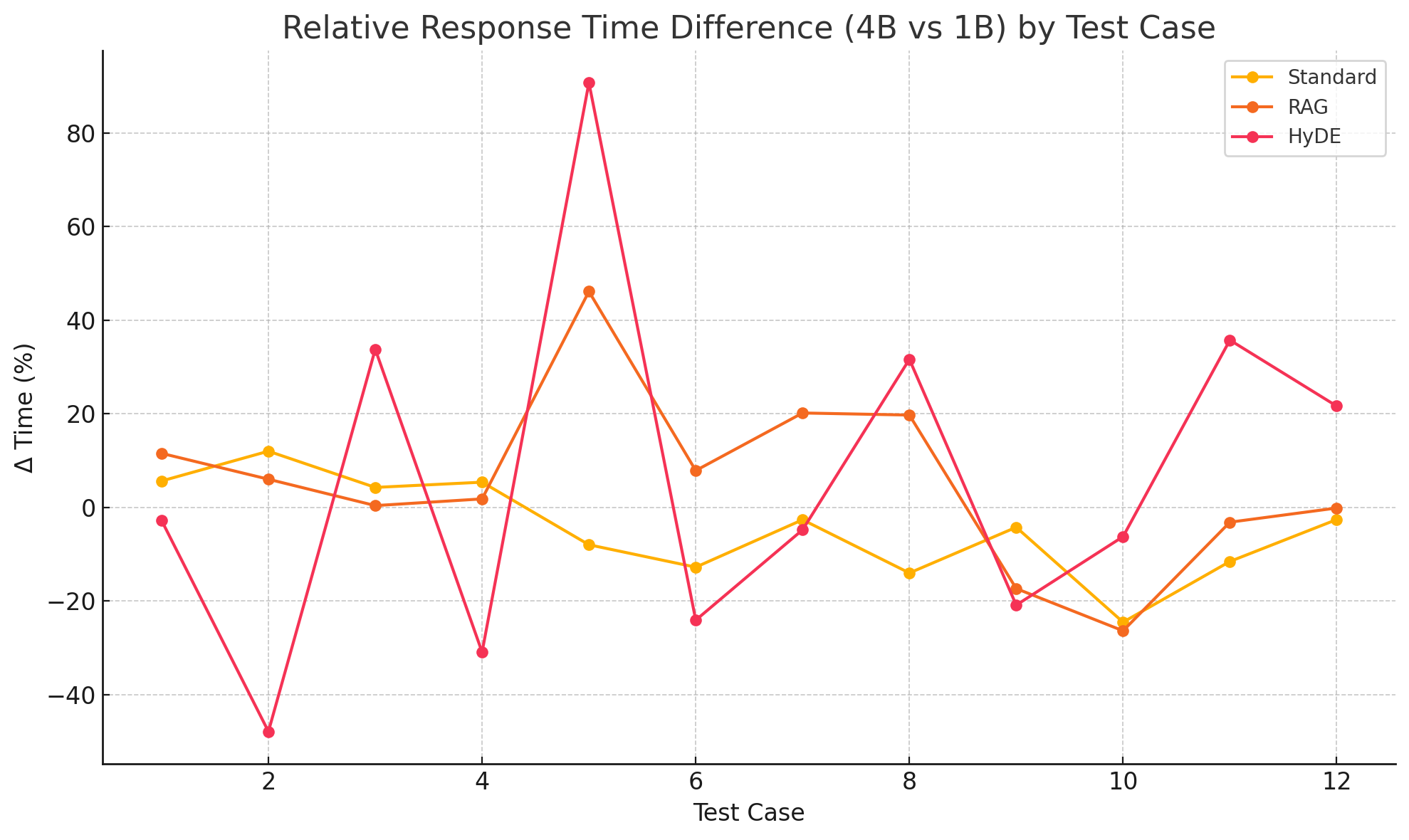}
    \caption{Relative Response Time Difference four and one B By Test Case}
    \label{fig:graph6}
\end{figure}

\cite{openai2023gpt4}

\end{document}